\documentclass[runningheads]{llncs}
\usepackage[T1]{fontenc}
\usepackage{graphicx,verbatim}
\usepackage{booktabs}
\usepackage{mathtools,amsmath,amssymb,amsfonts}
\usepackage{subcaption}
\usepackage{url}
\usepackage{cite}
\usepackage{xcolor}
\usepackage[hidelinks,breaklinks=true]{hyperref}

\newif\ifshowrevisions
\showrevisionsfalse 

\newcommand{\rev}[1]{%
  \ifshowrevisions
    \textcolor{red}{#1}%
  \else
    #1%
  \fi
}

\begin{document}
\title{Observation-Anchored Selective Assimilation for Longitudinal Tumor-State Proxy Forecasting in Post-Treatment Glioma}
\titlerunning{OASA for Longitudinal Tumor-State Proxy Forecasting}
%

\author{Yeonjae Jung\inst{1}
\and
Minwoo Shin\inst{1}\thanks{Corresponding author.}
}
\authorrunning{Y. Jung and M. Shin}
\institute{
Department of Software, Yonsei University, Wonju, 26493, Republic of Korea\\
\email{mshin@yonsei.ac.kr}
}
    
\maketitle              
\begin{abstract}
Post-treatment MRI in patients with glioma provides serial observations for updating patient-specific tumor-state proxy estimates, but variable appearances and trajectories complicate forecasting. We formulate forecasting as an observation-aware digital-twin update in which an intermediate observation anchors the patient-specific state. 
Among 203 patients and 594 follow-up time points, a predefined no-new-treatment criterion retained 120 of 236 candidate triplets, split into 81/24/15 training/validation/test triplets at the patient level.
Each time point was represented by a continuous voxel-wise tumor-state proxy map in \([0,1]\) derived from MRI lesion labels. A SegMamba-based single-step forecaster predicted update proposals from multimodal source-state tensors. Observation-Anchored Selective Assimilation (OASA) retained the observed intermediate proxy as the state anchor and selectively applied updates through a validation-selected tiered case-level rule and voxel-wise soft gate. We compared initial-scan forecasting, rollout without assimilation, latest-observation persistence, direct prediction, OASA, OASA + calibration, and morphological dilation. Checkpoints, OASA rules, and calibration thresholds were selected using validation data only. Across three seeds on 15 held-out test triplets, OASA maintained Dice at \(\tau=0.2\) comparable to persistence (\(0.6071 \pm 0.0025\) vs.\ \(0.6070\)) while yielding numerically higher Dice at \(\tau=0.5\) (\(0.4269 \pm 0.0079\) vs.\ \(0.3981\)), with a small RMSE increase. Calibration increased Dice at \(\tau=0.2\) to \(0.6178 \pm 0.0025\), increased false-positive (FP) support (\(11{,}836 \rightarrow 18{,}663\)), and reduced false-negative (FN) support (\(22{,}107 \rightarrow 17{,}536\)). This reflects near-threshold support calibration rather than improved biological predictive capability. Code is publicly available at \url{https://github.com/jsudg436/longitudinal-proxy-forecasting}.


\end{abstract}

\section{Introduction}
\rev{Longitudinal post-treatment MRI in patients with glioma} provides repeated patient-specific observations, but longitudinal tumor-state proxy forecasting cannot be fully addressed by static lesion segmentation or one-shot prediction alone~\cite{stupp2005rt_tmz,wen2010rano,wen2023rano2,hygino2011pseudoprogression}. Post-treatment appearances and proxy trajectories may be stable, increasing, decreasing, or non-monotonic, limiting a single fixed extrapolation rule. We represent each time point by a continuous voxel-wise tumor-state proxy map derived from MRI lesion labels and address longitudinal forecasting of this proxy.

Prior work on longitudinal tumor prediction, biophysical growth, reaction--diffusion modeling, and MRI synthesis motivates patient-specific forecasting~\cite{weizman2012prediction,swanson2003virtual,clatz2005simulation,konukoglu2010personalization,ezhov2023learnmorph,weidner2025learnableprior,predictgbm,tumorflow}. However, the newly acquired intermediate follow-up observation has received less attention as an explicit state-assimilation event in this setting. Directly predicting \(t_2\) from \(t_0\), rolling forecasts forward, or imposing fixed growth or trend priors does not explicitly use the observed \(t_1\) proxy as an updated patient-specific state from which subsequent forecasting should proceed.

We formulate this task as an observation-aware digital-twin update. In contrast to broader healthcare digital-twin work on patient-specific simulation and treatment modeling~\cite{hernandez2021digitaltwins,seo2026acoustic,cho2026dt}, we use the term to denote a state-estimation cycle that incorporates new follow-up information, rather than clinical treatment simulation or direct biological tumor-growth modeling. We introduce Observation-Anchored Selective Assimilation (OASA), which retains the observed intermediate proxy as the state anchor and selectively applies a model-predicted update proposal through a validation-selected tiered case-level gating rule and a voxel-wise soft change-probability gate. The single-step forecaster is implemented with a SegMamba-based 3D encoder--decoder~\cite{gu2023mamba,segmamba}; the contribution is the assimilation formulation rather than a claim of backbone superiority.

Our contributions are: (1) an observation-aware formulation for longitudinal tumor-state proxy forecasting in which the intermediate observation anchors the subsequent forecast; (2) OASA, a selective assimilation rule for model-predicted updates; and (3) \rev{an evaluation under patient-level data partitioning using continuous proxy error and proxy-support overlap at \(\tau=0.2\) and \(\tau=0.5\)}, together with a descriptive characterization of heterogeneous proxy trajectories. No checkpoint, OASA rule, calibration threshold, or final seed was selected based on held-out test performance.

\section{Method}
\begin{figure}[t]
\centering
\includegraphics[width=0.98\textwidth]{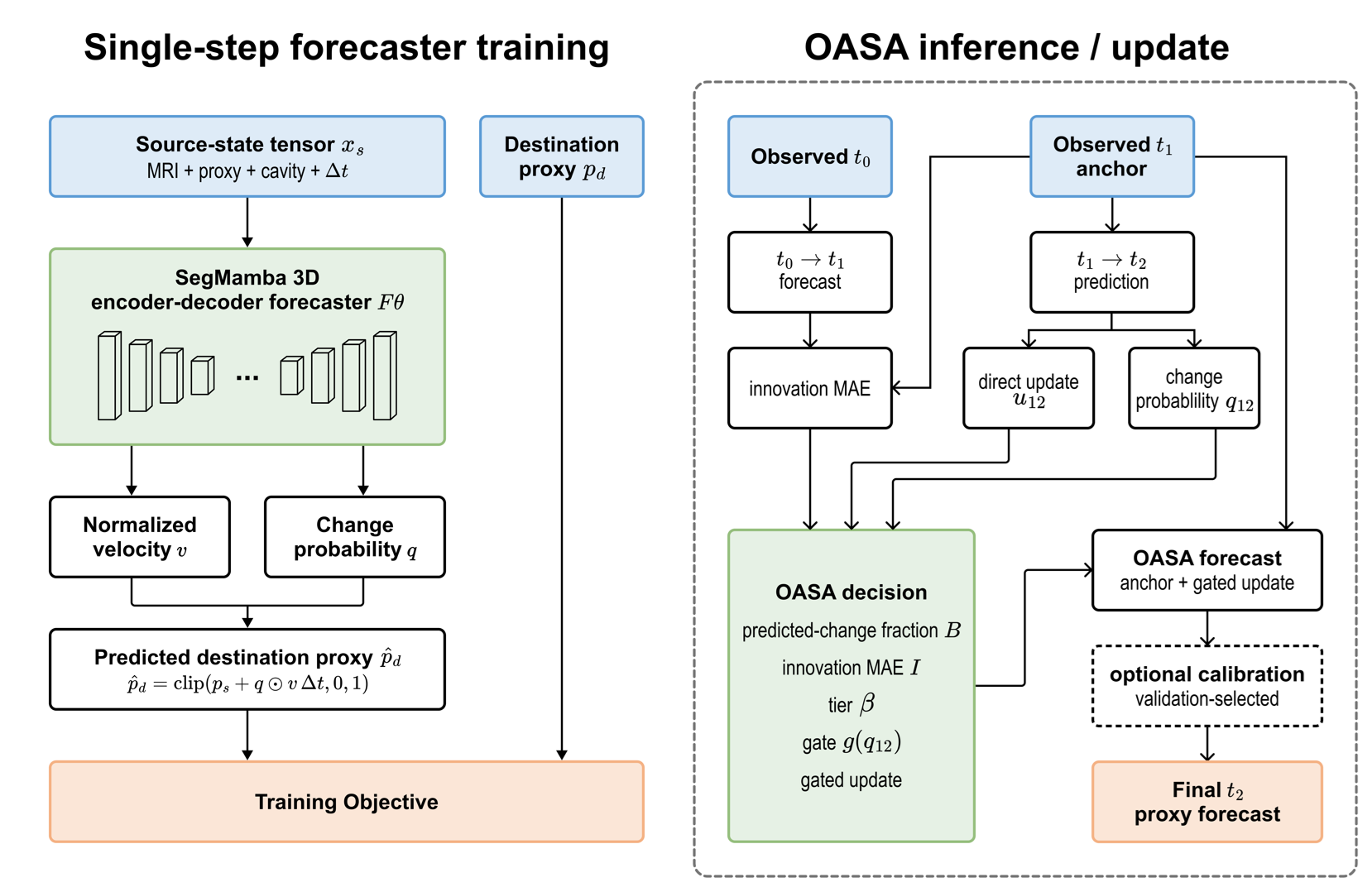}
\caption{OASA framework. A shared single-step forecaster predicts normalized velocity and voxel-wise change probability from a source-state tensor. The \(t_0\!\rightarrow\!t_1\) step estimates observation-innovation MAE, while the \(t_1\!\rightarrow\!t_2\) step generates a direct update proposal. OASA uses the predicted-change fraction, innovation MAE, and voxel-wise change probability to produce a gated update anchored to the observed \(t_1\) proxy. Optional validation-selected calibration adjusts near-threshold support.}
\label{fig:oasa_pipeline}
\end{figure}

Fig.~\ref{fig:oasa_pipeline} summarizes the proposed update cycle. A shared forecaster first produces single-step update proposals and change probabilities; OASA then uses the observed \(t_1\) proxy, the \(t_0\!\rightarrow\!t_1\) innovation, and the predicted-change burden to select a case-level update tier and apply voxel-wise gating, followed by optional near-threshold calibration.

\subsection{Tumor-State Proxy Forecasting}
Each forecasting triplet contains observed time points \((t_0,t_1)\) and future target \(t_2\). A tumor-state proxy map \(p_t\in[0,1]^{D\times H\times W}\), where \(D,H,W\) denote the three spatial dimensions, was generated at each time point \(t\) from glioma labels~\cite{menze2015brats,bakas2017tcga}. We assigned weights 0.7, 0.4, and 1.0 to non-enhancing tumor core, surrounding non-enhancing/FLAIR abnormality, and enhancing tumor labels, followed by Gaussian smoothing (\(\sigma=1.0\)), rescaling if the maximum exceeded 1, and clipping to \([0,1]\). Cavity voxels were encoded as a separate binary channel.

For a single forecast from source time point \(s\) to destination time point \(d\), the input is the 7-channel source-state tensor \(x_s=[m_s,p_s,c_s,\Delta t_{s\rightarrow d}]\), where brackets denote channel-wise concatenation. Here, \(m_s\) comprises four source MRI channels: contrast-enhanced T1-weighted (T1c), native T1-weighted (T1n), T2-weighted fluid-attenuated inversion recovery (T2F), and T2-weighted (T2W) MRI; \(p_s\) is the source tumor-state proxy; \(c_s\) is the binary source cavity map; and \(\Delta t_{s\rightarrow d}\) is a constant-valued interval channel containing the elapsed days from \(s\) to \(d\), normalized by 180. The model predicts the destination proxy \(p_d\) from \(x_s\), and does not receive \(t_0\) and \(t_1\) as a joint 14-channel input.

\subsection{Single-Step Forecaster}
Given \(x_s\), the SegMamba-based forecaster outputs a raw velocity channel \(r_v\) and a change-probability logit channel \(r_q\), converted to normalized velocity \(v=\tanh(r_v)\) and voxel-wise change probability \(q=\operatorname{sigmoid}(r_q)\). The direct forecast is
\begin{equation}
\hat p_d=\mathrm{clip}\left(p_s+q\odot v\odot\Delta t_{s\rightarrow d}\lambda_v,0,1\right),\qquad \lambda_v=1.0,
\label{eq:direct}
\end{equation}
where \(\hat p_d\) is the predicted destination proxy, \(\odot\) denotes element-wise multiplication, and \(\lambda_v\) is the velocity scale. Training used a forecast loss (\(L_1\)+Dice), velocity supervision on the effective velocity \(q\odot v\), binary cross-entropy for the change logit with binary target \(y_q=\mathbf{1}\{|p_d-p_s|\ge0.05\}\), and stable-delta regularization in voxels with \(|p_d-p_s|<0.05\), where \(\mathbf{1}\{\cdot\}\) denotes the indicator function. The velocity target was the time-normalized proxy change clipped to \([-1,1]\), and no auxiliary support head was active. The final objective was
\begin{equation}
\mathcal L=\mathcal L_{\mathrm{cwL1}}+0.5\mathcal L_{\mathrm{Dice}}+0.5\mathcal L_{\mathrm{vel}}+0.1\mathcal L_{\mathrm{BCE}}+\mathcal L_{\mathrm{stable}},
\label{eq:loss}
\end{equation}
where \(\mathcal L\) is the total training loss, and \(\mathcal L_{\mathrm{cwL1}}\), \(\mathcal L_{\mathrm{Dice}}\), \(\mathcal L_{\mathrm{vel}}\), \(\mathcal L_{\mathrm{BCE}}\), and \(\mathcal L_{\mathrm{stable}}\) denote change-weighted \(L_1\), soft Dice, velocity, change-logit binary cross-entropy, and stable-delta losses, respectively. The change-weighted terms assign a weight of \(1+2=3\) to voxels with proxy change of at least 0.05.

\subsection{Observation-Anchored Selective Assimilation}
OASA retains the observed intermediate proxy \(p_{t_1,\mathrm{obs}}\) as the state anchor. First, the \(t_0\!\rightarrow\!t_1\) forecast estimates observation-innovation MAE,
\begin{equation}
I=\mathrm{mean}|p_{t_1,\mathrm{obs}}-\hat p_{t_1|t_0}|.
\label{eq:innovation}
\end{equation}
Here, \(\hat p_{t_1|t_0}\) is the \(t_1\) proxy forecast from the \(t_0\) source state, and the mean is taken over all voxels.
Second, the \(t_1\!\rightarrow\!t_2\) step produces a direct proposal \(\hat p_{t_2,\mathrm{direct}}\) and change-probability map \(q_{12}\), where the subscript \(12\) denotes the \(t_1\!\rightarrow\!t_2\) interval. We define
\begin{equation}
u_{12}=\hat p_{t_2,\mathrm{direct}}-p_{t_1,\mathrm{obs}},\qquad
B=|\Omega|^{-1}\sum_{i\in\Omega}\mathbf{1}\{|u_{12}(i)|\ge0.05\}.
\label{eq:proposal_burden}
\end{equation}
where \(u_{12}\) is the proposed proxy update, \(\Omega\) is the voxel domain, \(|\Omega|\) is its voxel count, \(i\) indexes voxels, and \(B\) is the fraction of voxels with an absolute proposed update of at least 0.05.
The validation-selected rule evaluates high, mid, and low tiers in that order, using high when \(B\ge0.01\) and \(I\ge0.03\), mid when \(B\ge0.01\) and \(I\ge0.015\), low when \(B\ge0.005\), and persistence otherwise. 
\rev{The tier-specific factors \((\beta_{\mathrm{low}},\beta_{\mathrm{mid}}, \beta_{\mathrm{high}})=(0.1,0.1,1.0)\) control the fraction of the proposed update applied in the low, mid, and high update tiers, respectively; persistence applies no update. The tier labels denote branches of the validation-selected rule rather than calibrated confidence levels.}
In compact notation,
\begin{equation}
u_{\mathrm{OASA}}=\beta_{\mathrm{tier}}\,g(q_{12})\odot u_{12},\qquad
\hat p_{t_2,\mathrm{OASA}}=\mathrm{clip}(p_{t_1,\mathrm{obs}}+u_{\mathrm{OASA}},0,1).
\label{eq:oasa}
\end{equation}
Here, \(\beta_{\mathrm{tier}}\) is the factor for the selected tier, \(u_{\mathrm{OASA}}\) is the gated update, and \(\hat p_{t_2,\mathrm{OASA}}\) is the assimilated future proxy. The final voxel gate is soft, \(g(q_{12})=q_{12}\); 
\rev{thus OASA applies additional voxel-wise change-score weighting to an
already probability-weighted update proposal rather than to the raw
velocity field.}

\subsection{Near-Threshold Calibration}
OASA + calibration applies a validation-selected support adjustment:
\begin{equation}
\hat p_{\mathrm{cal}}(i)=
\begin{cases}
0.201, & \theta_{02}\le \hat p_{\mathrm{OASA}}(i)<0.2,\\
\hat p_{\mathrm{OASA}}(i), & \text{otherwise}.
\end{cases}
\label{eq:calibration}
\end{equation}
Here, \(\hat p_{\mathrm{cal}}\) is the calibrated proxy and \(\theta_{02}\) is the validation-selected lower eligibility threshold, set to 0.075 for all three seeds. Eligible voxels are raised to 0.201, immediately above the evaluation threshold of 0.2. This is interpreted as near-threshold support calibration, not evidence of improved biological predictive capability.

\section{Experiments}
\subsection{Data, Splits, and Metrics}
We used the de-identified MU-Glioma Post dataset, distributed through The Cancer Imaging Archive (TCIA) under a Creative Commons Attribution (CC BY) license~\cite{mahmoud2025mugliomapost,yaseen2025mugliomapost}. The original retrospective collection was approved by the University of Missouri Institutional Review Board (IRB \#2096253 MU), which granted a waiver of informed consent for the collection and sharing of de-identified data, as reported by the dataset authors~\cite{mahmoud2025mugliomapost}. \rev{According to the dataset-provided diagnosis labels, the source cohort comprised 203 patients, including 157 with glioblastoma and 46 with other glioma diagnoses.} We constructed 236 candidate triplets; the patient-level training, validation, and test splits contained 142/30/31 patients and 164/43/29 candidate triplets. A predefined no-new-treatment criterion excluded triplets with a recorded treatment start satisfying \(\mathrm{source\ day}<\mathrm{start\ day}\le\mathrm{destination\ day}\) in either \((t_0,t_1]\) or \((t_1,t_2]\). \rev{The filter retained 120 triplets from 72 patients, distributed as 81/24/15 training/validation/test triplets, for modeling and trajectory characterization. The retained cohort included 52 patients with glioblastoma and 20 with other glioma diagnoses; these groups contributed 90 and 30 triplets, respectively. Diagnosis labels were used only for cohort characterization and were not provided to the forecasting model.} These filtered trajectories should not be interpreted as untreated biological trajectories. MRI channels were standardized using nonzero voxels; training used foreground-centered \(128^3\) crops with jitter, while validation and test used deterministic centered crops.

We report RMSE for continuous proxy error and Dice at \(\tau=0.2\) and \(\tau=0.5\) for \rev{lower- and higher-threshold proxy-support overlap, respectively. The lower threshold captures broader proxy support, whereas the higher threshold evaluates a more restrictive region of the continuous proxy map}. Metrics were computed over all voxels in the centered \(128^3\) evaluation patch without an additional brain mask; FP@0.2 and FN@0.2 are voxel counts. Main learned results are mean \(\pm\) population standard deviation over three training seeds. For paired uncertainty analysis, metrics were first averaged across seeds within each triplet, followed by 10,000 paired triplet-level bootstrap resamples.

\subsection{Baselines and Validation Protocol}
We compared initial-scan forecasting, rollout without assimilation, persistence, direct prediction, OASA, OASA + calibration, and validation-selected radius-1 morphological dilation. Initial forecasting predicts \(t_2\) directly from \(t_0\); rollout predicts \(t_1\) from \(t_0\) and then propagates that prediction to \(t_2\); persistence copies the observed \(t_1\) proxy; and direct prediction forecasts \(t_2\) from the observed \(t_1\) state without OASA gating. Checkpoints, OASA gating rules, calibration threshold, and baseline parameters were selected on validation data only. The fixed patient-level split was shared across seeds; only initialization, training order, and sampling varied. The selected checkpoints corresponded to epochs 40, 25, and 45 for seeds 42, 43, and 44. Each validation-selected configuration was applied once to the 15 held-out test triplets, with no test-based seed or method selection. The public SegMamba implementation was trained for 60 epochs on two NVIDIA RTX A6000 GPUs using AdamW (learning rate \(1.1844{\times}10^{-4}\), weight decay \(7.1145{\times}10^{-4}\)), cosine scheduling without warmup, gradient clipping at 0.5, mixed precision, and batch size 2 per GPU. Online augmentation, dropout/drop path, and auxiliary support heads were disabled.

\section{Results}
\subsection{Main Comparison}
\begin{table}[t]
\centering
\caption{
\rev{Performance on 15 held-out test triplets from the fixed patient-level test split.}
Learned results are mean \(\pm\) population standard deviation over three seeds; deterministic baselines omit their zero seed standard deviations.
FP/FN are support \rev{voxel counts} at \(\tau=0.2\), reported in \(10^3\) voxels.
Bold indicates best.
}
\label{tab:main_comparison}
\footnotesize
\setlength{\tabcolsep}{3pt}
\begin{tabular}{@{}lccccc@{}}
\toprule
Method & RMSE\(\downarrow\) & Dice$_{0.2}\uparrow$ & Dice$_{0.5}\uparrow$ & FP$_{0.2}\downarrow$ & FN$_{0.2}\downarrow$ \\
\midrule
Initial
& \(0.0935{\pm}0.0025\)
& \(0.4952{\pm}0.0077\)
& \(0.2414{\pm}0.0067\)
& \(26.9{\pm}0.3\)
& \(24.4{\pm}1.1\) \\

Rollout
& \(0.0992{\pm}0.0046\)
& \(0.4700{\pm}0.0205\)
& \(0.2244{\pm}0.0134\)
& \(35.4{\pm}3.3\)
& \(23.0{\pm}0.9\) \\

Latest obs.
& \(0.0465\)
& \(0.6070\)
& \(0.3981\)
& \(\mathbf{11.2}\)
& \(22.1\) \\

Direct
& \(0.0674{\pm}0.0027\)
& \(0.6021{\pm}0.0189\)
& \(0.2777{\pm}0.0121\)
& \(16.3{\pm}1.7\)
& \(21.1{\pm}0.8\) \\

OASA
& \(0.0472{\pm}0.0005\)
& \(0.6071{\pm}0.0025\)
& \(\mathbf{0.4269{\pm}0.0079}\)
& \(11.8{\pm}0.3\)
& \(22.1{\pm}0.3\) \\

OASA + Cal.
& \(0.0471{\pm}0.0006\)
& \(\mathbf{0.6178{\pm}0.0025}\)
& \(\mathbf{0.4269{\pm}0.0079}\)
& \(18.7{\pm}0.5\)
& \(17.5{\pm}0.3\) \\

Dilation
& \(\mathbf{0.0464}\)
& \(0.6139\)
& \(0.3981\)
& \(22.9\)
& \(\mathbf{15.5}\) \\
\bottomrule
\end{tabular}
\end{table}

Table~\ref{tab:main_comparison} shows that initial-scan forecasting and rollout without assimilation had substantially lower overlap than methods using the intermediate observation. Latest-observation persistence was a strong conservative baseline. OASA maintained Dice at \(\tau=0.2\) comparable to persistence (\(0.6071\pm0.0025\) vs. \(0.6070\)) while yielding \rev{numerically} higher Dice at \(\tau=0.5\) (\(0.4269\pm0.0079\) vs. \(0.3981\)), with a small RMSE increase. 
\rev{Direct prediction had Dice at \(\tau=0.2\) similar to OASA and persistence, but lower Dice at \(\tau=0.5\) and higher RMSE, indicating that unconditional model-predicted updates were less favorable for proxy-support overlap at \(\tau=0.5\).}
OASA + calibration increased Dice at \(\tau=0.2\) to \(0.6178\pm0.0025\) without changing aggregate Dice at \(\tau=0.5\), but increased FP@0.2 from 11,836 to 18,663 while reducing FN@0.2 from 22,107 to 17,536. 
\rev{Morphological dilation was competitive in RMSE and Dice at \(\tau=0.2\), but had lower Dice at \(\tau=0.5\) and a higher FP voxel count at \(\tau=0.2\).}

A paired triplet-level bootstrap supported the main direct-update contrast: OASA increased Dice at \(\tau=0.5\) over direct prediction by 0.1493 (95\% CI 0.0291--0.3055) and reduced RMSE by 0.0202 (95\% CI 0.0113--0.0294). Confidence intervals (CIs) for OASA versus persistence and morphological dilation included zero for Dice metrics; calibration increased mean Dice at \(\tau=0.2\) by 0.0108 over OASA, but its CI included zero.

\subsection{Trade-off, Trajectories, and Qualitative Update}
\begin{figure}[t]
\centering
\begin{subfigure}[t]{0.48\textwidth}
\centering
\includegraphics[width=\linewidth]{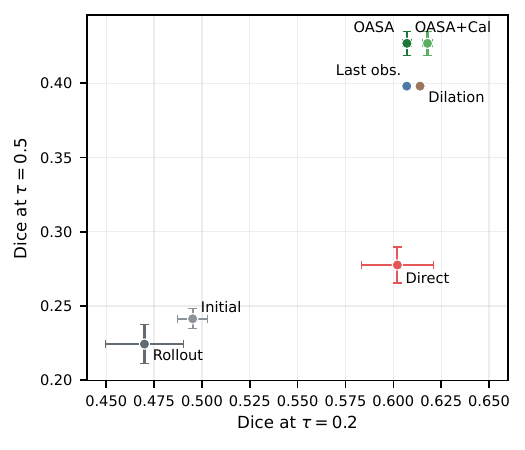}
\caption{\rev{Proxy-support overlap at \(\tau=0.2\) vs.\ \(\tau=0.5\).}}
\label{fig:support_core_tradeoff}
\end{subfigure}\hfill
\begin{subfigure}[t]{0.48\textwidth}
\centering
\includegraphics[width=\linewidth]{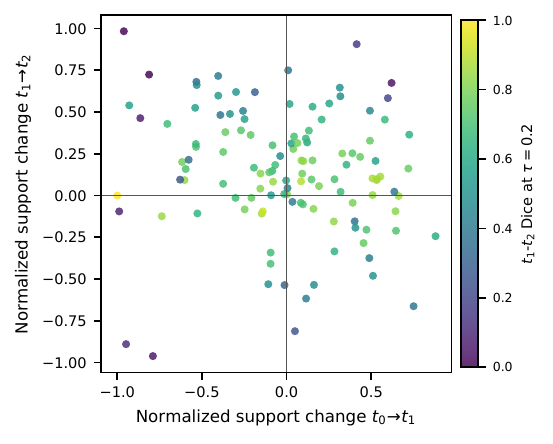}
\caption{Descriptive proxy-trajectory heterogeneity.}
\label{fig:proxy_trajectory_heterogeneity}
\end{subfigure}
\caption{Performance trade-off and descriptive support changes in the 120 no-new-treatment triplets. In (b), \(V_a\) and \(V_b\) are low-threshold support voxel counts at successive time points \(a\) and \(b\), and \(\Delta V_{ab}=(V_b-V_a)/\max(V_a,V_b,1)\); color denotes \(t_1\)--\(t_2\) Dice at \(\tau=0.2\). Trajectory categories were not used for model selection.}
\label{fig:tradeoff_trajectory}
\end{figure}

Fig.~\ref{fig:tradeoff_trajectory}(a) shows that direct prediction and morphological dilation can be competitive in \rev{Dice at \(\tau=0.2\) while showing lower Dice at \(\tau=0.5\)}. OASA and OASA + calibration \rev{shared the highest numerical Dice} at \(\tau=0.5\), whereas calibration shifted OASA toward higher Dice at \(\tau=0.2\) through near-threshold support adjustment. Fig.~\ref{fig:tradeoff_trajectory}(b) shows heterogeneous proxy trajectories. Patterns were non-monotonic (54/120, 45.0\%), mixed/mild (26/120, 21.7\%), increase-like (24/120, 20.0\%), stable-like (10/120, 8.3\%), or decrease-like (6/120, 5.0\%). This descriptive analysis motivates observation-aware updating rather than a single monotonic extrapolation rule.

\begin{figure}[t]
\centering
\includegraphics[width=0.80\textwidth]{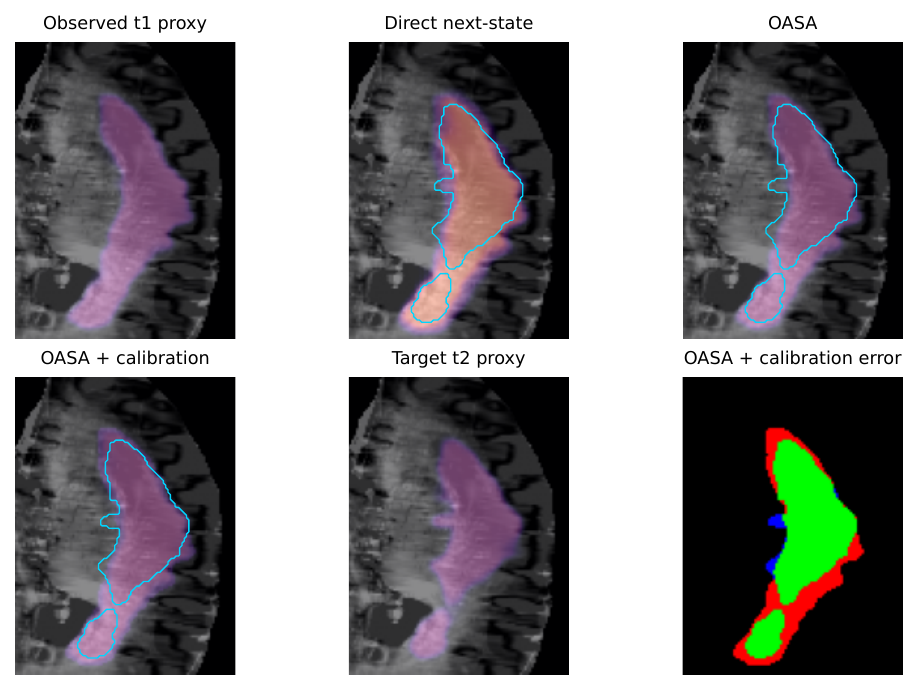}
\caption{\rev{Illustrative observation-anchored update for a held-out triplet. Among the six visualization candidates, the displayed case was selected because its Dice at \(\tau=0.2\) for OASA with calibration under seed 42 was closest to the median across the candidate set. The case was selected only for visualization after all model, checkpoint, rule, threshold, and seed choices had been fixed. Cyan contours denote future target support at \(\tau=0.2\). The support-comparison map uses green, red, and blue to denote true-positive (TP), false-positive (FP), and false-negative (FN) voxels, respectively, at the same threshold.}}
\label{fig:visual_panel}
\end{figure}

Fig.~\ref{fig:visual_panel} illustrates that direct prediction, OASA, and OASA + calibration produce different support patterns for the same future target proxy. OASA retains the observed intermediate proxy as the state anchor and applies a conservative gated update rather than unconditionally applying the model-predicted update proposal. Near-threshold calibration is shown as a secondary support adjustment.

\section{Discussion and Conclusion}
\rev{Our results support observation anchoring as a conservative alternative to unconditional direct prediction. OASA was comparable to persistence in Dice at \(\tau=0.2\) while showing numerically higher Dice at \(\tau=0.5\), with a small RMSE increase relative to persistence. Direct prediction showed greater across-seed variability than OASA for all reported metrics. Morphological dilation occupied a different operating point: compared with OASA, it showed numerically higher Dice and a higher FP voxel count at \(\tau=0.2\), but lower Dice at \(\tau=0.5\).}

\rev{OASA + calibration reduced FN and increased Dice at \(\tau=0.2\), but also increased FP. By construction, it left Dice at \(\tau=0.5\) unchanged. This operation should be interpreted as a near-threshold support adjustment rather than improved biological prediction; uncalibrated OASA remains the primary method.}

No-new-treatment proxy trajectories were frequently non-monotonic, motivating the use of intermediate observations. Although recorded treatment starts were excluded, these should not be considered untreated biological trajectories. Growth and reaction--diffusion frameworks~\cite{predictgbm,tumorflow} motivate extrapolation, but we neither reproduce them nor claim superiority. SegMamba provides the 3D backbone; the contribution is observation-anchored assimilation.

This study forecasts a label-derived proxy rather than clinical progression, recurrence, treatment response, or biological tumor burden. Results from one institution and 15 held-out test triplets should be interpreted descriptively. Future work should evaluate larger cohorts from multiple institutions and clinical endpoints. Overall, OASA provides a conservative strategy for observation-aware forecasting.

\begin{credits}
\subsubsection{\ackname}

This work was supported by the Korea Medical Device Development Foundation grant funded by the Korean government (the Ministry of Science and ICT, the Ministry of Trade, Industry and Energy, the Ministry of Health and Welfare, and the Ministry of Food and Drug Safety) (Grant No. RS-2026-25543484).

This research was also supported by the ANCHOR Program through the Gangwon ANCHOR Center, funded by the Ministry of Education (MOE) and Gangwon State (G.S.), Republic of Korea (Grant No. 2026-ANCHOR-10-006).

This research was further supported by the Ministry of Science and ICT (MSIT), Korea, under the National Program in Medical AI Semiconductor (Grant No. 2024-0-00096), supervised by the Institute of Information \& Communications Technology Planning \& Evaluation (IITP) in 2026.

\subsubsection{\discintname}
The authors have no competing interests to declare that are relevant to the content of this article.

\end{credits}
%

\bibliographystyle{splncs04}
\bibliography{references}
\end{document}